%% file: main.tex
\documentclass{article}
\usepackage{microtype}
\usepackage{graphicx}
\usepackage{subfigure}
\usepackage{booktabs} 
\usepackage{amsopn} 
\usepackage{mathtools}
\usepackage{amsmath, amssymb}
\usepackage{bbm}
\usepackage{stmaryrd}
\usepackage{enumitem}
\usepackage{xspace}

\newcommand{\doi}[1]{}

\DeclareMathOperator*{\argmin}{arg\,min}
\DeclareMathOperator*{\argmax}{arg\,max}
\DeclarePairedDelimiter\norm{\lVert}{\rVert}%

\newcommand{\parag}[1]{\medskip \noindent \textbf{{#1}.}}
\newcommand{\pomdp}{\textsc{Pomdp}\xspace}

\usepackage[accepted]{icml2018}

\icmltitlerunning{Programmatically Interpretable Reinforcement Learning}

\newcommand{\rl}{\textsc{Rl}\xspace}
\newcommand{\pirl}{\textsc{Pirl}\xspace}
\newcommand{\sygurl}{\textsc{Pirl}\xspace}

\newcommand{\algo}{\textsc{Ndps}\xspace}
\newcommand{\torcs}{\textsc{Torcs}\xspace}
\newcommand{\drl}{\textsc{Drl}\xspace}

\newcommand{\figlabel}[1]{\label{fig:#1}}
\newcommand{\figref}[1]{Figure~\ref{fig:#1}}

\begin{document}

\twocolumn[
\icmltitle{Programmatically Interpretable Reinforcement Learning}

\begin{icmlauthorlist}
\icmlauthor{Abhinav Verma}{rice}
\icmlauthor{Vijayaraghavan Murali}{rice}
\icmlauthor{Rishabh Singh}{gb}
\icmlauthor{Pushmeet Kohli}{deepmind}
\icmlauthor{Swarat Chaudhuri}{rice}
\end{icmlauthorlist}

\icmlaffiliation{rice}{Rice University}
\icmlaffiliation{gb}{Google Brain}
\icmlaffiliation{deepmind}{Deepmind}

\icmlcorrespondingauthor{Abhinav Verma}{averma@rice.edu}

\icmlkeywords{Reinforcement Learning, Program Synthesis}

\vskip 0.3in
]

\printAffiliationsAndNotice{}

\input{abs}

\input{sec_intro}

\input{sec_problem}

\input{sec_algo}

\input{sec_environs}

\input{sec_eval}

\input{sec_related}

\input{sec_conclusion}

\section*{Acknowledgements}

This research was partially supported by NSF Award CCF-1162076 and DARPA MUSE Award \#FA8750-14-2-0270.

\bibliography{ogsyn-bibfile}
\bibliographystyle{icml2018}

\newpage
\twocolumn[
\icmltitle{Appendix: Programmatically Interpretable Reinforcement Learning}
]
\input{sec_appendix}

\end{document}

%% file: abs.tex
\begin{abstract}
We present a reinforcement learning framework, called {\em Programmatically
    Interpretable Reinforcement Learning} (\pirl), that is designed to 
generate interpretable and verifiable
  agent policies. Unlike the popular Deep
  Reinforcement Learning (\drl) paradigm, which represents policies by
  neural networks, \pirl 
represents policies using a high-level, domain-specific programming language. Such 
  programmatic policies have the benefits of being more easily
  interpreted than neural networks, and being amenable to verification by symbolic
  methods. We propose a new method, called {\em
    Neurally Directed Program Search} (\algo), for solving the
  challenging nonsmooth optimization problem of finding a
  programmatic policy with maximal reward. \algo works by first learning a neural policy
  network using \drl, and then performing a local search over
  programmatic policies that
  seeks to minimize a distance from this neural
  ``oracle''. We evaluate \algo on the task of learning to drive a
  simulated car in the \torcs car-racing environment. We demonstrate
  that \algo is able to discover human-readable policies that pass
  some significant performance bars. We also show that
\pirl policies can have 
  smoother trajectories, and can be 
  more easily transferred to environments not encountered during
  training, than corresponding policies discovered by \drl.
\end{abstract}

%% file: sec_intro.tex
\section{Introduction}\label{sec:intro}

Deep reinforcement learning (\drl) has had a massive impact on the
field of machine learning and has led to remarkable successes in the
solution of many challenging
tasks~\citep{mnih2015human,AlphaGo,silver2017mastering}.  While neural
networks have been shown to be very effective in learning good
policies, the expressivity of these models makes them difficult to
interpret or to be checked for consistency for some desired
properties, and casts a cloud over the use of such representations in
safety-critical applications.

Motivated to overcome this problem, we propose a learning framework,
called {\em Programmatically Interpretable Reinforcement Learning}
(\pirl)\footnote{\pirl is pronounced Pi-R-L (as in $\pi$-\rl)}, that is based on the idea of learning policies that are
represented in a human-readable language.  The \pirl framework is
parameterized on a high-level {\em programming language} for policies.
A problem instance in \pirl is similar to a one in traditional \rl,
but also includes a {\em (policy) sketch} that syntactically defines a set of programmatic policies
in this language.  The objective is to find a program in this set with
maximal long-term reward.

Intuitively, the policy programming language and the sketch characterize what we
consider ``interpretable''. In addition to interpretability, the
syntactic restriction on policies has three key benefits. First, the language
can be used to implicitly encode the learner's inductive bias that will
be used for generalization. Second, the language can allow effective
pruning of undesired policies to make the search for a good policy
more efficient. Finally, it 
allows us to use symbolic program verification techniques to formally reason
about the learned policies and check consistency with correctness
properties. At the same time, policies in \pirl can have rich 
semantics, for example allowing actions to depend on events far back
in history. 

A key technical challenge in \pirl is that the space of policies
permitted in an instance can be vast and nonsmooth, making
optimization extremely challenging. To address this, we propose a new
algorithm called {\em Neurally Directed Program Synthesis}
(\algo). The algorithm first uses \drl to compute a neural policy
network that has high performance, but may not be expressible in the
policy language. This network is then used to direct a local search
over programmatic policies. In each iteration of this search, we
maintain a set of ``interesting'' inputs, and update the program so as
to minimize the distance between its outputs and the outputs of the
neural policy (an ``oracle'') on these inputs. The set of interesting
inputs is updated as the search progresses. This strategy, inspired by
imitation learning~\cite{dagger,schaal1999imitation}, allows us to
perform direct policy search in a highly nonsmooth policy space.

We evaluate our approach in the task of learning to drive a simulated
car in the \torcs car-racing environment~\citep{TORCS}, as well as
three classic control games (we discuss the former in the main paper,
and the latter in the Appendix). Experiments demonstrate that \algo is
able to find interpretable policies that, while not as performant as
the policies computed by \drl, pass some significant performance
bars. Specifically, in \torcs, our policy sketch allows an unbounded
set of programs with branches guarded by unknown conditions, each
branch representing a Proportional-Integral-Derivative (PID)
controller~\citep{aastrom1995pid} with unknown parameters. The policy we
obtain can successfully
complete a lap of the race, and the use of the neural oracle is
key to doing so. Our results also suggest that a well-designed sketch can serve as a
regularizer. Due to constraints imposed by the sketch, the
policies for \torcs that \algo learns lead to smoother trajectories
than the corresponding neural policies, and can tolerate greater noise. The policies are also more easily transferred to new
domains, in particular race tracks not seen during training. Finally,
we show, using several properties, that the programmatic
policies that we discover are amenable to verification using
off-the-shelf symbolic techniques.

%% file: sec_problem.tex
\section{Programmatically Interpretable Reinforcement Learning}
\label{sec:sygurl}

In this section, we formalize the problem of programmatically interpretable reinforcement learning (\sygurl). 

\newcommand{\Cals}{\mathcal{S}}
\newcommand{\A}{\mathcal{A}}
\newcommand{\ra}{\rightarrow}
\newcommand{\T}{\mathcal{T}}
\newcommand{\Lang}{\mathcal{L}}
\newcommand{\sem}[1]{[\![#1]\!]}
\newcommand{\Prob}{\mathbf{Pr}}
\newcommand{\Expec}{\mathbf{E}}
\newcommand{\Policies}{\mathcal{P}}
\newcommand{\Observ}{\mathcal{O}}
\newcommand{\Init}{\mathit{Init}}
\newcommand{\Op}{\mathit{Ops}}
\renewcommand{\S}{\mathcal{S}}

\newcommand{\combinec}{\mathbf{combine}}
\newcommand{\op}{\oplus}

\newcommand{\mapc}{\mathbf{map}}
\newcommand{\filterc}{\mathbf{filter}}
\newcommand{\headc}{\mathbf{hd}}
\newcommand{\tailc}{\mathbf{tl}}
\newcommand{\pickc}{\mathbf{peek}}

\newcommand{\hole}{\chi}

\newcommand{\ifc}{\mathbf{if}}
\newcommand{\letc}{\mathbf{let}}
\newcommand{\thenc}{\mathbf{then}}
\newcommand{\pushc}{\mathbf{push}}
\newcommand{\alphac}{x}

\newcommand{\elsec}{\mathbf{else}}
\newcommand{\foldc}{\mathbf{fold}}
\newcommand{\orc}{\mathbf{or}}
\newcommand{\andc}{\mathbf{and}}
\newcommand{\Action}{\mathit{Action}}
\newcommand{\Obs}{\mathit{Obs}}
\newcommand{\Accel}{\mathit{Accel}}
\newcommand{\latest}{\mathtt{Latest}}
\newcommand{\previous}{\mathtt{Previous}}
\newcommand{\rpm}{\mathtt{RPM}}
\newcommand{\tangle}{\mathtt{TrackPos}}
\newcommand{\Param}{\mathit{Param}}

We model a reinforcement learning setting as a {\em Partially Observable Markov Decision Process (\pomdp)} $M = (\Cals, \A, \Observ, T(\cdot | s, a), Z(\cdot | s), r, \Init, \gamma)$. Here, $\Cals$ is the set of {(environment) states}. $\A$ is the set of {\em actions} that the learning agent can perform, and $\Observ$ is the set of {\em observations} about the current state that the agent can make. An agent action $a$ at the state $s$ causes the environment state to change probabilistically, and the destination state follows the distribution $T(\cdot | s, a)$. The probability that the agent makes an observation $o$ at state $s$ is $Z(o | s)$. The {\em reward} that the agent receives on performing action $a$ in state $s$ is given by $r(s, a)$.  $\Init$ is the {initial distribution} over environment states.  Finally, $0 < \gamma < 1$ is a real constant that is used to define the agent's aggregate reward over time.

A {\em history} of $M$ is a sequence $h = o_0, a_0, \dots, a_{k-1}, o_k$, where $o_i$ and $a_i$ are, respectively, the agent's observation and action at the $i$-th time step. Let $\mathcal{H}_M$ be the set of histories in $M$. A {\em policy} is a function $\pi: \mathcal{H}_M \ra \A$ that maps each history as above to an action $a_k$.
For each policy, we can define a set of histories that are possible when the agent follows $\pi$. We assume a mechanism to simulate the POMDP and sample histories that are possible under a policy.  
The policy also induces a distribution over possible rewards $R_i$ that the agent receives at the $i$-th time step. 
The agent's {\em expected aggregate reward} under $\pi$ is given by $R(\pi) = \Expec[\sum_{i = 0}^\infty \gamma^i R_i]$. The goal in reinforcement learning is to discover a policy $\pi^*$ that maximizes $R(\pi)$. 

\parag{A Programming Language for Policies} The distinctive feature of
\sygurl is that policies here are expressed in a high-level, domain-specific programming language. 
Such a language can be defined in many ways. However, to facilitate search through the space of programs expressible 
in the language, it is desirable for the language to express computations as compactly and canonically as possible. Because of this, we propose to express parameterized policies using a functional language based on a small number of side-effect-free combinators. It is known from prior work on program synthesis~\cite{DBLP:conf/pldi/FeserCD15} that such languages offer natural advantages in program synthesis.

We collectively refer to observations and actions, as well as auxiliary integers and reals generated during computation, as {\em atoms}. Our language considers two kinds of data: atoms and sequences of atoms (including histories). We assume a finite set of {\em basic operators} over atoms that is rich enough to capture all common operations on observations and actions. 
  
\figref{language} shows the syntax of this language. The nonterminals $E$ and $\alphac$ represent expressions that evaluate to atoms and histories, respectively. We sketch the semantics of the various language constructs below. 
\begin{itemize}
\item $c$ ranges over a universe of numerical constants, and $\op$ is a basic operator;
\item $\pickc~(\alphac, i)$ returns the observation from the $i$-th time step from a history $\alphac$, and $\pickc~(\alphac, -1)$ is used as shorthand for the most recent observation;
\item $\foldc$ is a standard higher-order combinators over sequences with the semantics: \\
 $\foldc(f, [e_1, \ldots, e_k], e) = f(e_k, f(e_{k-1}, ... f(e_1, e)))$
\item $x, x_1, x_2$ are variables.
As usual, unbound variables are assumed to be inputs. 
\end{itemize}
The language comes with a type system that distinguishes between different types of atoms, and ensures that language constructs are used consistently. The type system can catch common errors, such as applying $\pickc~(\alphac, k+1)$ to a history of size $k$. This type system identifies a set of expressions whose inputs are histories and outputs are actions. These expressions are known as {\em programmatic policies}, or simply {\em programs}. 

The combinator over sequences, $\foldc$, will be operating over histories in our programs. Since histories can be of variable lengths, we restrict these combinators to operate on only the last $n$ elements of a sequence, for some fixed $n$. This restriction provides us the ability to use these combinators in a consistent manner.

\begin{figure}
{\small
\begin{eqnarray*}
E & ::= & c \mid x \mid \op(E_1,\dots, E_k) \mid \pickc~(\alphac, i) \mid \\& & 
\foldc~((\lambda x_1, x_2.~E_1),\alpha) 
\\
\end{eqnarray*}
}
\vspace{-0.3in}
\caption{Syntax of the policy language.}\figlabel{language}
\label{fig:syntax}
\vspace{-0.1in}
\end{figure}

\paragraph{Sketches.} 

Discovering an optimal programmatic policy from the vast space of legitimate programs is typically impractical without some prior on the shape of target policies. \sygurl allows the specification of such priors through instance-specific syntactic models called {\em sketches}. 

We define a sketch as a grammar of expressions over atoms and
sequences of atoms. The sketch places restrictions on the kinds of basic operators that will be considered during policy search. Formally, the sketch is obtained by restricting the grammar in
\figref{language}. The set of programs permitted by a
sketch $\S$ is denoted by $\sem{\S}$.

\paragraph{\sygurl.}

The \sygurl problem can now be stated as follows. Suppose we are given a \pomdp $M$ and a sketch $\S$. Our goal is to find a {program} $e^* \in \sem{\S}$ 
with optimal reward:
\begin{eqnarray}
\label{eq:synthesis}
e^* &= &  \argmax_{e \in \sem{\S}} R(e).
\end{eqnarray}

\paragraph{Example.}

Now we consider a concrete example of \sygurl, considered in more detail in  Section \ref{sec:evaluation}. 

Suppose our goal is to make a (simulated) car complete laps on a track. We want to do so by learning policies for tasks like steering and acceleration.
Suppose we know that we could get well-behaved policies by using PID control --- specifically, by switching back and forth between a set of PID controllers. However, we do not know the parameters of these controllers, and neither do we know the conditions under which we should switch from one controller to another.
We can express this knowledge using the following sketch:
\begin{eqnarray*}
P & ::= & \pickc((\epsilon - h_i), -1) \\
I & ::= & \foldc(+, \epsilon - h_i) \\
D & ::= & \pickc(h_i, -2)- \pickc(h_i, -1) \\
C &::=& c_1 * P + c_2 * I + c_3 * D \\
B & ::=  & c_0 + c_1 * \pickc(h_1, -1) + \dots  \\
& & \dots + c_k * \pickc(h_m, -1) > 0 \mid \\
& & \qquad B_1 ~\orc~ B_2 \mid B_1 ~\andc~ B_2 \\
E  & ::= & C \mid \ifc~B~\thenc~E_1~\elsec~E_2. 
\end{eqnarray*}

Here, $E$ represents programs permitted by the sketch. The program's
input is a history $h$. We assume that this sequence is split into a
set of sequences $\{h_1, \dots, h_m\}$,  where $h_i$ is the sequence
of observations from the $i$-th of $m$ sensors. The sensor's most recent reading is given by $\pickc(h_k, -1)$, and its second most recent reading is $\pickc(h_k, -2)$. 
The operators $+$, $-$, $*$, $>$, and if-then-else are as usual. 
The program (optionally) evaluates a set of boolean conditions  ($B$)
over the current sensor readings, then chooses among a set of discretized PID
controllers, represented by $C$. 
In the definition of $C$, $P$ is the proportional term, $I$ is the discretized
integral term (calculated via a $\foldc$), and $D$ is a
finite-difference approximation of the derivative term, $\epsilon$ is a known constant and represents a fixed target for the controller, $(\epsilon - h_i)$ performs an element-wise operation on the sequence $h_i$. The symbols $c_i$ are real-valued parameters. Recall that the $\foldc$ acts over a fixed-sized window on the history, and hence can be used as a discrete approximation of the integral term in a PID controller.

The program in Figure~\ref{fig:code} shows the body of a policy for
acceleration that the \algo algorithm finds given this sketch in the
\torcs car racing environment. The program's input consists of histories for 29 sensors; however, only two of them, $\tangle$ and $\rpm$, are actually used in the program. While the sensor $\tangle$ (for the position of the car relative to the track axis) is used to decide which controller to use, only the $\rpm$ sensor is needed to calculate the acceleration. 

\begin{figure*}[ht]
	\vskip 0.1in
	$$
	\begin{array}{l}
	\ifc~(0.001 - \pickc(h_\tangle, -1) > 0 )~\andc~(0.001 + \pickc(h_\tangle, -1) > 0) \\
	\quad \thenc~ 3.97 * \pickc((0.44 - h_\rpm), -1) +  0.01 * \foldc(+, (0.44 - h_\rpm)) + 48.79 * ( \pickc(h_\rpm, -2)- \pickc(h_\rpm, -1)) \\
	\quad \elsec~ \;\, 3.97 * \pickc((0.40 - h_\rpm), -1) + 0.01 * \foldc(+, (0.40 - h_\rpm)) +  48.79* ( \pickc(h_\rpm, -2)- \pickc(h_\rpm, -1))
	\end{array}
	$$
	\caption{A programmatic policy for acceleration, automatically discovered by the \algo algorithm. $h_\rpm$ and $h_\tangle$ represent histories for the $\rpm$ and $\tangle$ sensors, respectively.
	}
	\label{fig:code}
\end{figure*}

%% file: sec_algo.tex
\newcommand{\NN}{\mathcal{N}}
\newcommand{\syn}{\mathit{syn}}

\section{Neurally Directed Program Search}
\label{sec:algorithm}

\paragraph{Imitating a Neural Policy Oracle.}

The \algo algorithm is a direct policy search that is guided by a neural ``oracle''. Searching over policies is a standard approach in reinforcement learning. However, the nonsmoothness of the space of programmatic policies poses a fundamental challenge 
to the use of such an approach in \sygurl. For example, a conceivable way of solving the search problem would be to 
define a neighborhood relation over programs and perform local search. However, in practice, the objective $R(e)$ of such a search can vary irregularly, leading to poor performance (see Section~\ref{sec:evaluation} for experimental results on this).

In contrast, \algo starts by using \drl to compute a neural policy oracle $e_\NN$ for the given environment. This policy is an approximation of the programmatic policy that we seek to find. To a first approximation, \algo is a local search over programmatic policies that seeks to find a program $e^*$ that closely imitates the behavior of $e_\NN$. The main intuition here is that distance from $e_\NN$ is a simpler objective than the reward function $R(e)$, which aggregates rewards over a lengthy time horizon. This approach can be seen to be a form of imitation learning~\citep{schaal1999imitation}.

The {\em distance} between $e_\NN$ and the estimate $e$ of $e^*$ in a search iteration is defined as 
$
d(e_\NN, e) = \sum_{h \in \mathcal{H}} \norm{e(h) - e_\NN(h)}, 
$
where $\mathcal{H}$ is a set of ``interesting'' inputs (histories) and $\norm{\cdot}$ is a suitable norm. During the iteration, we search the neighborhood of $e$ for a program $e'$ that minimizes this distance. At the end of the iteration, 
$e'$ becomes the new estimate for $e^*$.

\parag{Input Augmentation} One challenge in the algorithm is that under the policy $e$, the agent may encounter histories that are not possible under $e_\NN$, or any of the programs encountered in previous iterations of the search. For example, while searching for a steering controller, we may arrive at a program that, under certain conditions, steers the car into a wall, an illegal behavior that the neural policy does not exhibit. Such histories would be irrelevant to the distance between $e_\NN$ and $e$ if the set $\mathcal{H}$ were constructed ahead of time by simulating $e_\NN$, and never updated. This would be unfortunate as these are precisely the inputs on which the programmatic policy needs guidance. 

Our solution to this problem is {\em input augmentation}, or periodic updates to the set $\mathcal{H}$.  More precisely, after a certain number of search steps for a fixed set $\mathcal{H}$, and after choosing the best available synthesized program for this set, we sample a set of additional histories by simulating the current programmatic policy, and add these samples to $\mathcal{H}$.

\subsection{Algorithm Details} 
\label{algodetails}

\begin{algorithm}[h]
	\caption{Neurally Directed Program Search}
	\label{alg:ogSynthesis}
{
	\begin{algorithmic}
		\STATE {\bfseries Input:} POMDP $M$, neural policy $e_{\NN}$, sketch $\S$
		\STATE $ \mathcal{H} \gets \mathtt{create\_histories}(e_{\NN}, M)$
		\STATE $e \gets \mathtt{initialize}(e_{\NN}, \mathcal{H}, M, \S)$
		\STATE $R \gets  \mathtt{collect\_reward}(e, M)$
		\REPEAT
		\STATE $ (e^\prime, R^\prime) \gets (e, R) $ 
		\STATE $\mathcal{H} \gets \mathtt{update\_histories}(e, e_\NN, M, \mathcal{H})$
		\STATE $\mathcal{E} \gets \mathtt{neighborhood\_pool}(e)$
		\STATE $e \gets  \argmin_{e' \in \mathcal{E}} \sum_{h \in \mathcal{H}} \norm{e'(h) - e_\NN(h)}$ 
		\STATE $R \gets  \mathtt{collect\_reward}(e, M)$
	    \UNTIL{$R^\prime \geq R$}  
		\STATE {\bfseries Output:} $ e^\prime$		
	\end{algorithmic}
}
\end{algorithm}

We show pseudocode for \algo in Algorithm~\ref{alg:ogSynthesis}. 
The inputs to the algorithm are a \pomdp $M$, a neural policy $e_\NN$ for $M$ that serves as an oracle, and a sketch $\S$. The algorithm first samples a set of histories of $e_\NN$ using the procedure $\mathtt{create\_histories}$. Next it uses the routine $\mathtt{initialize}$ to generate the program that is the starting point of the policy search. Then the procedure $\mathtt{collect\_reward}$ calculates the expected aggregate reward $R(e)$ (described in Section~\ref{sec:sygurl}), by simulating the program in the \pomdp. 

From this point on, \algo iteratively updates its estimate $e$ of the target program, as well as its estimate $\mathcal{H}$ of the set of interesting inputs used for distance computation. To do the former, \algo uses the procedure $\mathtt{neighborhood\_pool}$ to generate a space of programs that are structurally similar to $e$, then finds the program in this space that minimizes distance from $e_\NN$. The latter task is done by the routine $\mathtt{update\_histories}$, which heuristically picks interesting inputs in the trajectory of the learned program and then obtains the corresponding actions from the oracle for those inputs. This process goes on until the iterative search fails to improve the estimated reward $R$ of $e$.

The subroutines used in the above description can be implemented in many ways. Now we elaborate on our implementation of the important subroutines of \algo. 

\paragraph{The optimization step.}
The search for a program $e'$ at minimal distance from the neural oracle can be implemented in many ways. The approach we use has two steps. First, we enumerate a set of {\em program templates} --- numerically parameterized programs --- that are structurally similar to $e$ and are permitted by the sketch $\S$, giving priority to shorter templates. Next, we find optimal parameters for the enumerated templates.
Our primary tool for the second step is Bayesian optimization \cite{BayesianOpt}, though we also explored a symbolic optimization technique based on Satisfiability Modulo Theories (SMT) solving  (Appendix~\ref{sec:opt}).

\paragraph{The initialization step.}
The performance of \algo turns out to be quite sensitive to the choice of the program that is the starting point of the search. 
Our initialization routine $\mathtt{initialize}$ is broadly similar to the optimization step, in that it attempts to find programs that closely imitate the oracle through a combination of template enumeration and parameter optimization. 
However, rather than settling on a single program, $\mathtt{initialize}$ generates a pool of programs that are close in behavior to the oracle. After this, it simulates the programs in the POMDP and returns the program that achieves the highest reward.

%% file: sec_environs.tex
\section{Environments for Experiments}
\label{sec:environments}
In this section, we describe the environments (modeled by POMDPs) on which we evaluated the \algo algorithm. 

\paragraph{\torcs.} We use \algo to generate controllers for cars in \emph{The Open Racing Car Simulator} (\torcs) \cite{TORCS}. \torcs has been used extensively in AI research, for example in \cite{TORCS:FuzzyControl}, \cite{TORCS:NN}, and \cite{TORCS:Overtake} among others. \cite{NN:DDPG} has shown that a Deep Deterministic Policy Gradient (DDPG) network can be used in \rl environments with continuous action spaces. The \drl agents for \torcs in this paper implement this approach.

In its full generality \torcs provides a rich environment with input from up to $89$ sensors, and optionally the 3D graphic from a chosen camera angle in the race. The controllers have to decide the values of 5 parameters during game play, which correspond to the acceleration, brake, clutch, gear and steering of the car. 
Apart from the immediate challenge of driving the car on the track, controllers also have to make race-level strategy decisions, like making pit-stops for fuel. A lower level of complexity is provided in the \emph{Practice Mode} setting of TORCS. In this mode all race-level strategies are removed. Currently, so far as we know, state-of-the-art \drl models are capable of racing only in \emph{Practice Mode}, and this is also the environment that we use. Here we consider the input from $29$ sensors, and decide values for the acceleration and steering actions.

The sketches used in our experiments are as in the example in Section~\ref{sec:sygurl}, and provide the basic structure of a proportional-integral-derivative (PID) program, with appropriate holes for parameter and observation values. To obtain a practical implementation, we constrain the fold calculation to the five latest observations of the history. This constraint corresponds to the standard strategy of automatic (integral) error reset in discretized PID controllers \cite{PIDControl}.

Each track in \torcs can we viewed as a distinct \pomdp. In our implementation of \algo for \torcs we choose one track and synthesize a program for it. Whenever the algorithm needs to interact with the \pomdp, we use the program or \drl agent to race on the track. For example, in the procedure $\mathtt{collect\_reward}$ we use the synthesized program to race one lap, and the reward is a function of the speed, angle and position of the car at each time step.

For the $\mathtt{create\_histories}$ procedure we use the \drl agent to complete one lap of the track (an {\em episode}), recording the sensor values and environment state at each time step. The $\mathtt{update\_histories}$ procedure uses a two step process. First, the synthesized program is used to race one lap and we store the sequence of observations (given by sensor values) $o_1,o_2,\dots$ provided by \torcs during this lap. Then, we use the \drl agent to generate the corresponding action $a_i$ for each observation $o_i$. Each tuple $(o_i, a_i)$ is then added to the set of histories.

\paragraph{Classic Control Games.} 
In addition to \torcs, we evaluated our approach in three classic control games, \emph{Acrobot}, {\em CartPole}, and {\em MountainCar}. These games provide simpler \rl environments, with fewer input sensors than \torcs and only a single discrete action at each time step, compared to two continuous actions in \torcs. These results appear in Appendix~\ref{apn:evaluations}.

%% file: sec_eval.tex
\section{Experimental Analysis}
\label{sec:evaluation}
Now we present an empirical evaluation of the effectiveness of our algorithm in solving the \sygurl problem. We synthesize programs for two \torcs tracks, CG-Speedway-1 and Aalborg. These tracks provide varying levels of difficulty, with Aalborg being the more difficult track of the two.

\subsection{Evaluating Performance}
A controller's performance is measured according to two metrics, \textit{lap time} and \textit{reward}. To calculate the lap time, the programs are allowed to complete a three lap race, and we report the average time taken to complete a lap during this race. The reward function is calculated using the car's velocity, angle with the track axis, and distance from the track axis. The same function is used to train the DRL agent initially. In the experiments we compare the average reward per time step, obtained by the various programs. 

We compare among the following \rl agents:
\begin{enumerate}[label=A\arabic*:]
	\item \textsc{DRL}. An agent which uses \drl to find a policy
	represented as a deep neural
	network. The specific \drl algorithm we use is Deep
	Deterministic Policy Gradients~\citep{ddpg}, which has
	previously been used on \torcs.  
	\item $\textit{Naive}$. Program synthesized without access to a policy oracle. 
	\item $\textit{NoAug}$. Program synthesized without input augmentation.
	\item $\textit{NoSketch}$. Program synthesized in our policy language without sketch guidance.
	\item $\textit{NoIF}$. Programs permitted by a restriction of
          our sketch that does not permit conditional branching.
	\item \algo. The Program generated by the \algo algorithm.
\end{enumerate}

In Table~\ref{table:performance} we present the performance results of the above list. The lap times in that table are given in minutes and seconds. The \textsc{Timeout} entries indicate that the synthesis process did not return a program that could complete the race, within the specified timeout of twelve hours.

These results justify the various choices that we made in our \algo algorithm architecture, as discussed in Section~\ref{sec:algorithm}. In many cases those choices were necessary to be able to synthesize a program that could successfully complete a race. As a consequence of these results, we only consider the DRL agent and the \algo program for subsequent comparisons.

The $\textit{NoAug}$ and $\textit{NoSketch}$ agents are unable to
generate programs that complete a single lap on either track. In the
case of $\textit{NoSketch}$ this is because the syntax of the policy
language (Figure~\ref{fig:syntax}), defines a very large program
space. If we randomly sample from this space without any constraints
(like those provided by the sketch), then the probability of getting a
good program is extremely low and hence we are unable to reliably
generate a program that can complete a lap. The $\textit{NoAug}$ agent
performs poorly because without input augmentation, the synthesizer
has no guidance from the oracle regarding the ``correct'' behavior once the program deviates even slightly from the oracle's trajectory.

The \algo algorithm is biased towards generating simpler programs to aid in interpretability. In the \algo algorithm experiments we allow the synthesizer to produce policies with up to five nested $\ifc$ statements. However, if two policies have \textsc{Lap Times} within one second of each other, then the algorithm chooses the one with fewer $\ifc$ statements as the output. This is a reasonable choice because a difference of less than one second in \textsc{Lap Times} can be the result of different starting positions in the \torcs simulator, and hence the performance of such policies is essentially equivalent.

\begin{table}[h]
	\caption{Performance results in \torcs. Lap time is given in Minutes:Seconds. Timeout indicates that the synthesizer did not return a program that completed the race within the specified timeout.}
	\label{table:performance}
	\begin{center}
		\begin{small}
			\begin{sc}
				\begin{tabular}{l c c cc}
					\toprule
					Model &  \multicolumn{2}{ c }{CG-Speedway-1} &  \multicolumn{2}{ c }{Aalborg}  \\
					\cline{2-5}
					& Lap Time & Reward  & Lap Time & Reward   \\
					\midrule					
					Drl & 54.27 & 118.39 & 1:49.66 &  71.23  \\
					$\textit{Naive}$  &  2:07.09 & 58.72 & Timeout  & $-$\\
					$\textit{NoAug}$  & Timeout  & $-$ & Timeout  & $-$\\
					$\textit{NoSketch}$ & Timeout  & $-$ & Timeout  & $-$\\
					$\textit{NoIF}$ & 1:01.60  & 115.25 & 2:45.13  & 52.81  \\
					\algo & 1:01.56 & 115.32 &  2:38.87 & 54.91\\	
					\bottomrule
				\end{tabular}
			\end{sc}
		\end{small}
	\end{center}
\end{table}

\subsection{Qualitative Analysis of the Programmatic Policy}
We provide qualitative analysis of the inferred programmatic policy through the lens of interpretability, and its behavior in acting in the environment.

\paragraph{Interpretability.} 
Interpretability is a qualitative metric, and cannot be easily
demonstrated via  experiments. The \drl policies are considered uninterpretable because their policies
are encoded in black box neural networks. 
In contrast, the \pirl policies are compact and
human-readable by construction, as
exemplified by the acceleration policy in Figure~\ref{fig:code}. More examples of our synthesized policies are given in Appendix~\ref{sec:examples}.

\paragraph{Behavior of Policy.}
Our experimental validation showed that the programmatic policy was less aggressive in terms of its use of actions and resulting in smoother steering actions. Numerically, we measure smoothness in Table~\ref{table:smoothness} by comparing the population standard deviation of the set of steering actions taken by the program during the entire race. In Figure~\ref{fig:smoothness} we present a scatter plot of the steering actions taken by the DRL agent and the \algo program during a slice of the CG-Speedway-1 race. As we can see, the \algo program takes much more conservative actions.

\begin{figure}[t]
	\vspace{-0.1in}
	\begin{center}
		\centerline{\includegraphics[scale=0.4]{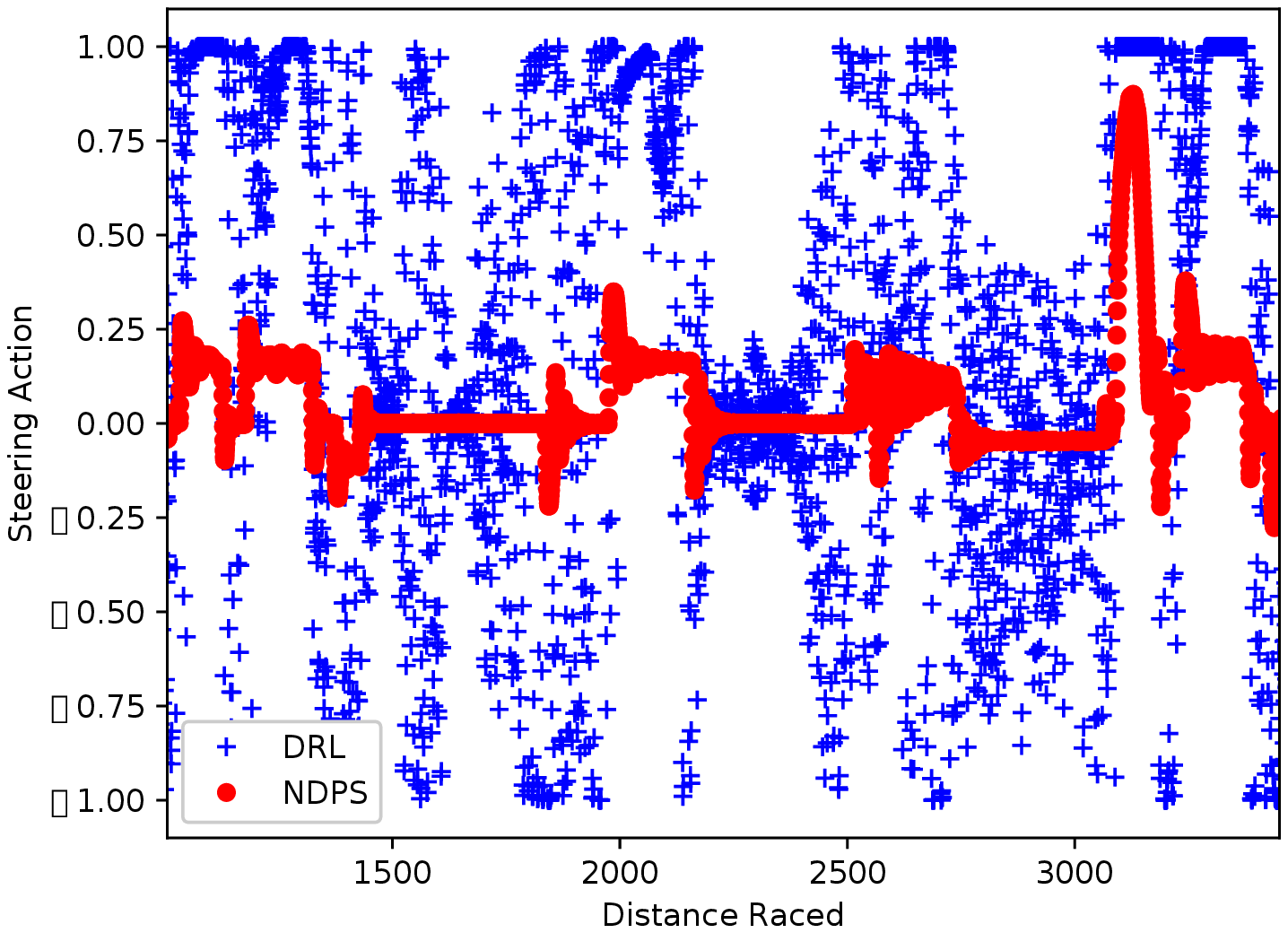}}
		\caption{Slice of steering actions taken by the DRL and \algo agents, during the CG-Speedway-1 race. This figure demonstrates that the \algo agent drives more smoothly.}
		\label{fig:smoothness}
	\end{center}
	\vspace{-0.2in}
\end{figure}

\begin{table}[h]
	\caption{Smoothness measure of agents in \torcs, given by the standard deviation of the steering actions during a complete race. Lower values indicate smoother steering.}
	\label{table:smoothness}
	\begin{center}
		\begin{small}
			\begin{sc}
				\begin{tabular}{l c c }
					\toprule
					Model &  CG-Speedway-1 & Aalborg  \\ 
					\midrule
					Drl & 0.5981  & 0.9008  \\
					\algo & 0.1312 & 0.2483  \\					
					\bottomrule
				\end{tabular}
			\end{sc}
		\end{small}
	\end{center}
\end{table}

\subsection{Robustness to Missing/Noisy Features}
To evaluate the robustness of the agents with respect to defective sensors we introduce a \emph{Partial Observability} variant of \torcs. In this variant, a random sample of $j$ sensors are declared defective. During the race, one or more of these defective sensors are blocked with some fixed probability. Hence, during game-play, the sensor either returns the correct reading or a \emph{null} reading. For sufficiently high block probabilities, both agents will fail to complete the race. In Table~\ref{table:partial} we show the distances raced for two values of the block probability, and in Figure~\ref{fig:partial} we plot the distance raced as we increase the block probability on the Aalborg track. In both these experiments, the set of defective sensors was taken to be $\{\rpm, \tangle\}$ because we know that the synthesized programs crucially depend on these sensors.

\begin{table}[h]
	\caption{Partial observability results in \torcs blocking sensors $\{\rpm, \tangle\}$ . For each track and block probability we give the distance, in meters, raced by the program before crashing.}
	\label{table:partial}
	\begin{center}
		\begin{small}
			\begin{sc}
				\begin{tabular}{l c c cc}
					\toprule
					Model &  \multicolumn{2}{ c }{CG-Speedway-1} &  \multicolumn{2}{ c }{Aalborg}  \\ 
					\cline{2-5}
					& 50\% & 90\%  & 50\% & 90\%     \\
					\midrule				
					Drl & 21 & 17 & 71 &  20  \\					
					\algo & 1976 & 200 &  1477	 & 287\\
					\bottomrule
				\end{tabular}
			\end{sc}
		\end{small}
	\end{center}
\end{table}

\begin{figure}[h]
	\vskip -0.11in
	\begin{center}
		\centerline{\includegraphics[width=\columnwidth]{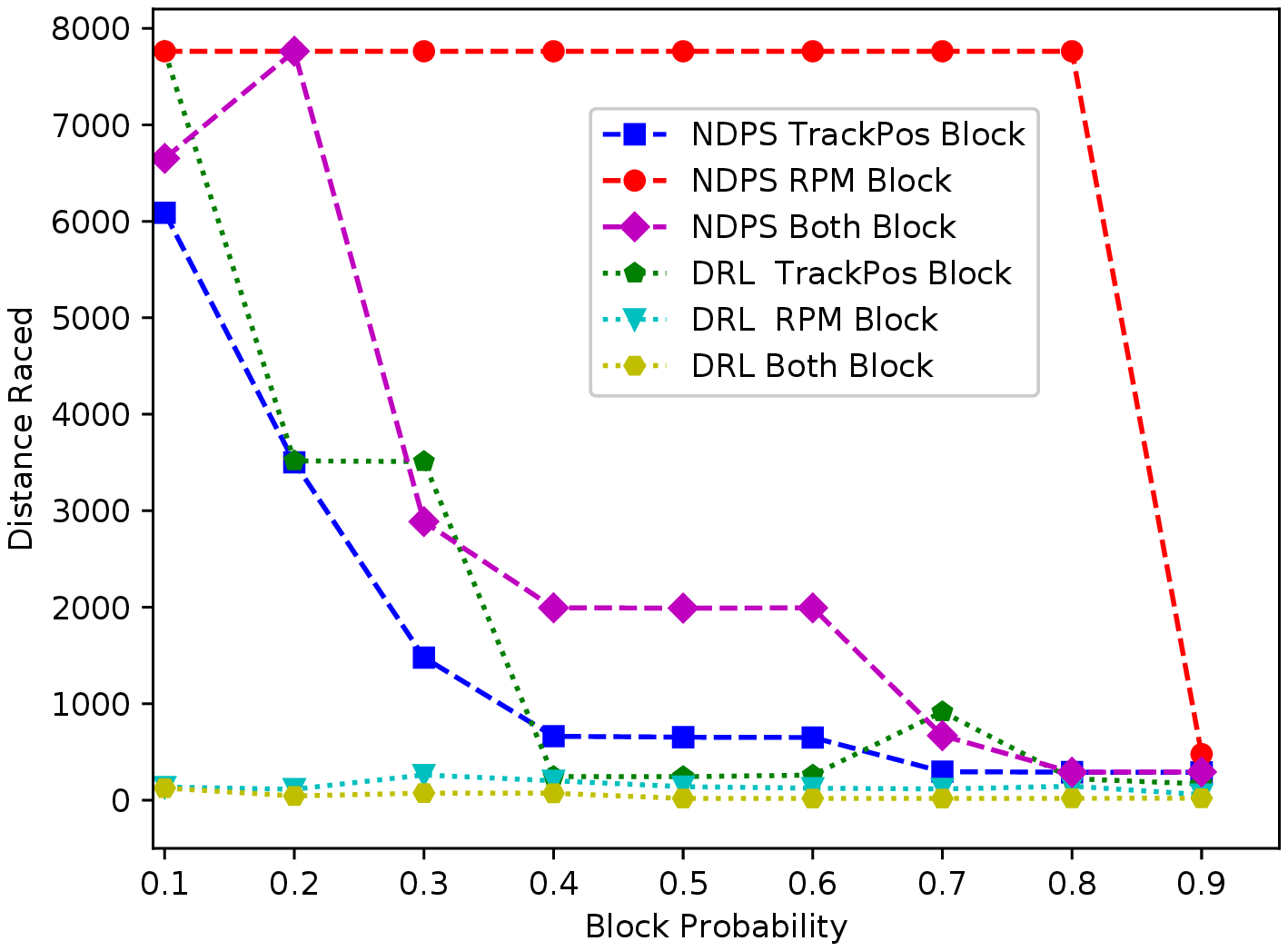}}
		\caption{Distance raced by the agents as the block probability increases for a particular sensor(s) on Aalborg. The \algo agent is more robust to blocked sensors.}
		\label{fig:partial}
	\end{center}
	\vskip -0.2in
\end{figure}

\subsection{Evaluating Generalization to New Instances}
To compare the ability of the agents to perform on unseen tracks, we executed the learned policies on tracks of comparable difficulty. For agents trained on the CG-Speedway-1 track, we chose CG track 2 and E-Road as the transfer tracks, and for Aalborg trained tracks we chose Alpine 2 and Ruudskogen. As can be seen in Tables \ref{table:cgspeed} and \ref{table:aalborg}, the \algo programmatically synthesized program far outperforms the DRL agent on unseen tracks. The DRL agent is unable to complete the race on any of these transfer tracks. This demonstrates the transferability of the policies \algo finds.

\begin{table}[h]
	\caption{Transfer results with training on CG-Speedway-1. `Cr'
		indicates that the agent crashed after racing the specified distance.}
	\label{table:cgspeed}
	\begin{center}
		\begin{small}
			\begin{sc}
				\begin{tabular}{l c c c c}
					\toprule
					Model &  \multicolumn{2}{ c }{CG track 2}  &  \multicolumn{2}{ c }{E-Road}   \\ 
					\cline{2-5}
					& Lap Time & Reward & Lap Time & Reward   \\
					\midrule					
					DRL &  Cr 1608m & $-$&  Cr 1902m & $-$ \\					
					\algo & 1:40.57 & 110.18 & 1:51.59 & 98.21 \\					
					\bottomrule
				\end{tabular}
			\end{sc}
		\end{small}
	\end{center}
	\caption{Transfer results with training on Aalborg. `Cr' denotes the agent crashed, after racing the specified distance.}
	\label{table:aalborg}
	\begin{center}
		\begin{small}
			\begin{sc}
				\begin{tabular}{l c c c c }
					\toprule
					Model &  \multicolumn{2}{ c }{Alpine 2}  &  \multicolumn{2}{ c }{Ruudskogen}   \\ 
					\cline{2-5}
					& Lap Time & Reward & Lap Time & Reward   \\
					\midrule					
					DRL &   Cr 1688m  & $-$&  Cr 3232m & $-$ \\					
					\algo & 3:16.68 &  67.49 & 3:19.77 & 57.69 \\										
					\bottomrule
				\end{tabular}
			\end{sc}
		\end{small}
	\end{center}
	\vspace{-0.2in}
\end{table}

\subsection{Verifiability of Policies}
Now we use established symbolic verification techniques to automatically prove two properties of policies generated by \algo. So far as we know, the current state of the art neural network verifiers cannot verify the DRL network we are using in a reasonable amount of time, due to the size and complexity of the network used to implement the DDPG algorithm. For example, the Reluplex~\cite{reluplex} algorithm was tested on networks at most 300 nodes wide, whereas our network has three layers with 600 nodes each, and other smaller layers.

\paragraph{Smoothness Property} For the program given in Figure~\ref{fig:code} we proved, we have $  \forall k, \ \sum_{i=k}^{k+5}\norm{\pickc(h_\rpm, i+1) - \pickc(h_\rpm, i)} < 0.006  \implies  \norm{\pickc(h_{\mathtt{Accel}}, k+1) - \pickc(h_{\mathtt{Accel}}, k)} < 0.49 $. Intuitively, the above logical implication means that if the sum of the consecutive differences of the last six $\rpm$ sensor values is less than $0.006$, then the acceleration actions calculated at the last and penultimate step will not differ by more than $0.49$. Similarly, for a policy given in Appendix~\ref{sec:examples}, we prove $  \forall k, \; \sum_{i=k}^{k+5}\norm{\pickc(h_\tangle, i+1) - \pickc(h_\tangle, i)} < 0.006 \implies \norm{\pickc(h_{\mathtt{Steer}}, k+1) - \pickc(h_{\mathtt{Steer}}, k)} < 0.11$. This proof gives us a guarantee of the type of smooth steering behavior that we empirically examined earlier in this section. 

\paragraph{Universal Bounds} We can prove that the program in Figure~\ref{fig:code} satisfies the property $  \forall i \; (0 \leq \pickc(h_\rpm, i) \leq 1 \wedge -1 \leq \pickc(h_\tangle, i) \leq 1) \implies (\norm{\pickc(h_\mathtt{Steer}, i)} < 101.08 \wedge -54.53 < \pickc(h_\mathtt{Accel}, i) < 53.03)$.  Intuitively, this means that we have proved global bounds for the action values in this environment, assuming reasonable bounds on some of the input values. In the \torcs environment these bounds are not very useful, since the simulator clips these actions to certain pre-specified ranges. However, this experiment demonstrates that our framework allows us to prove universal bounds on the actions, and this could be a critical property for other environments.

%% file: sec_related.tex
\section{Related Work}
\label{sec:related}

\paragraph{Syntax-Guided Synthesis.} 
The original formulation of inductive program synthesis is to search for a program in a hypothesis space (programming language) that is consistent with a specification (such as IO examples). However, this search is often intractable because of the large (potentially infinite) hypothesis space. One of the key ideas to make this search tractable is to provide the synthesizer a sketch of the desired program in addition to the examples, for example in \cite{Sketch:Main} and \cite{DBLP:conf/pldi/FeserCD15}. The program sketch in addition to providing structure to the search space also allows users to provide additional insights. This approach has been generalized in a framework called Syntax-Guided Synthesis (\textsc{Sygus})~\cite{sygus}. Our $\sygurl$ approach is inspired by \textsc{Sygus} in the sense that we also use a high-level grammar to constrain the shape of the possible learnt policies in a policy language grammar. However, unlike \textsc{Sygus} and previous sketch-based synthesis approaches that use logical constraints as specification, \sygurl searches for policies with quantitative objectives.
\paragraph{Imitation Learning.} Imitation learning~\cite{schaal1999imitation} has been a successful paradigm for reducing the sample complexity of reinforcement learning algorithms by allowing the agent to leverage the additional supervision provided in terms of expert demonstrations for the desired behaviors. The \textsc{DAgger} (Dataset Aggregation) algorithm~\cite{dagger} is an iterative algorithm for imitation learning that learns stationary deterministic policies, where in each iteration $i$ it uses the current learnt policy $\pi_i$ to collect new trajectories and adds them to the dataset $D$ of all previously found trajectories. The policy for the next iteration $\pi_{i+1}$ is a policy that best mimics the expert policy $\pi^*$ on the whole dataset $D$. Our Neurally Directed Program Search (\algo) is inspired by the \textsc{DAgger} algorithm, where we use the trained DeepRL agent as the expert (oracle), and iteratively perform IO augmentation for unseen input states explored by our synthesized policy with the current best reward. However, one key difference is that \algo uses the expert trajectories to only guide the local program search in our policy language grammar to find a policy with highest rewards, unlike the imitation learning setting where the goal is to match the expert demonstrations perfectly.
\paragraph{Neural Program Synthesis and Induction.} Many recent efforts use neural networks for learning programs. These efforts have two flavors. In \emph{neural program induction}, the goal is to learn a network that encodes the program semantics using internal weights. These architectures typically augment neural networks with differentiable computational substrates such as memory (Neural Turing Machines~\cite{ntm}), modules (Neural RAM~\cite{nram}) or data-structures such as stacks~\cite{stackrnn}, and formulate the program learning problem in an end-to-end differentiable manner. In \emph{neural program synthesis}, the architectures generate programs directly as outputs using multi-task transfer learning (e.g. \textsc{RobustFill}~\cite{RobustFill}, \textsc{DeepCoder}~\cite{DeepCoder}, \textsc{Bayou}~\cite{Bayou}), where the network weights are used to guide the program search in a DSL. There have also been some recent approaches to use \rl for learning to search programs in DSLs~\cite{rlkarel,rlbf}. Our approach falls in the category of program synthesis approaches where we synthesize policies in a policy language. However, we learn richer policy programs with continuous parameters using the \algo algorithm.
\paragraph{Interpretable Machine Learning.} Many recent efforts in deep learning aim to make deep networks more interpretable \citep{NN:Understanding,NN:Mythos, SymbolicRL, SymbRL2, SymbRL3, SymbRL4}. There are three key approaches explored for interpreting DNNs: i) generate input prototypes in the input domain that are representatives of the learned concept in the abstract domain of the top-level of a DNN, ii) explaining DNN decisions by relevance propagation and computing corresponding representative concepts in the input domain, and iii) Using symbolic techniques to explain and interpret a DNN. 
Our work differs from these approaches in that we are replacing the DRL model with human readable source code, that is programmatically synthesized to mimic the policy found by the neural network. Working at this level of abstraction provides a method to apply existing synthesis techniques to the problem of making DRL models interpretable.
\paragraph{Verification of Deep Neural Networks.} Reluplex~\cite{reluplex} is an SMT solver that supports linear real arithmetic with ReLU constraints, and has been used to verify several properties of DNN-based airborne collision avoidance systems, such as not producing erroneous alerts and uniformity of alert regions. Unlike Reluplex, our framework generates interpretable program source code as output, where we can use traditional symbolic program verification techniques~\cite{king1976symbolic} to prove program properties.

%% file: sec_conclusion.tex
\section{Conclusion}
\label{sec:conclusion}

We have introduced a framework for interpretable reinforcement learning, called \pirl. Here, policies are represented in a high-level language. The goal is to find a policy that fits a syntactic ``sketch'' and also has optimal long-term reward. We have given an algorithm inspired by imitation learning, called \algo, to achieve this goal. Our results show that the method is able to generate interpretable policies that clear reasonable performance goals, are amenable to symbolic verification, and, assuming a well-designed sketch, are robust and easily transferred to unseen environments.

The experiments in this paper only considered environments with symbolic inputs. Handling perceptual inputs may raise additional algorithmic challenges, and is a natural next step. Also, in this paper, we only considered deterministic (if memoryful) policies. Extending our framework to stochastic policies is a goal for future work. 
Finally, while we explored policies in the context of reinforcement learning, one could define similar frameworks for other learning settings.

%% file: sec_appendix.tex
\appendix
\section{Evaluation on Classic Control Games}
\label{apn:evaluations}
In this section, we provide results of additional experimental evaluation on some classic control games. We use the OpenAI Gym environment implementation of these games. A brief description of these games is given below.

We used the {\sc Duel-DDQN} algorithm~\citep{duel} to obtain our neural policy oracle for these games, rather than \textsc{Ddpg}, as an implementation of Duel-DDQN already appears on the OpenAI Gym leader-board.  

\begin{table}[h]
	\caption{Rewards achieved in Classic Control Games. Acrobot does not have threshold at which it is considered solved.}
	\label{table:openai}
	\begin{center}
		\begin{small}
			\begin{sc}
				\begin{tabular}{l c c c}
					\toprule
					& Acrobot & CartPole & MountainCar  \\
					\midrule
                          \textsc{Solved} & $-$ & 195 & -110  \\
                                        \drl & -63.17 & 197.53 & -84.73  \\
					\algo-SMT & -84.16  &  183.15 &  -108.06 \\
                                        \algo-BOPT & -127.21 & 143.21 & -143.86 \\
                                                    \textsc{Minimum} & -200 & 8 & -200  \\
					\bottomrule
				\end{tabular}
			\end{sc}
		\end{small}
	\end{center}
\end{table}

\paragraph{Acrobot.} This environment consists of a two link, two joint robot. The joint between the links is actuated. At the start of the episode, the links are hanging downwards. At every timestep the agent chooses an action that correspond to applying a force to move the actuated link to the right, to the left, or to not applying a force. The episode is over once the end of the lower link swings above a certain height. The goal is to end the episode in the fewest possible timesteps. 

We use the OpenAI Gym `Acrobot-v1' environment. This implementation is based on the system presented in~\citep{JMLR:v16:geramifard15a}. Each observation is a set consisting of readings from six sensors, corresponding to the rotational joint angles and velocities of joints and links. The action space is discrete with three elements, and at each timestep the environment returns the observation and a reward of $-1$. An episode is terminated after 200 time steps irrespective of the state of the robot. This is an unsolved environment, which means it does not have a specified reward threshold at which it's considered solved.

\paragraph{CartPole.} This environment consists of a pole attached by an un-actuated joint to a cart that moves along a frictionless track. At the beginning, the pole is balanced vertically on the cart. The episode ends when the pole is more than $15^\circ$ from vertical, or the cart moves more than 2.4 units from the center.  At every timestep the agent chooses to apply a force to move the cart to the right or to the left, and the goal is to prevent an episode from ending for the maximum possible timesteps. 

We use the OpenAI Gym `CartPole-v0' environment, based on the system presented in~\citep{Barto83}. The sensor values correspond to the cart position, cart velocity, pole angle and pole velocity. The action space is discrete with two elements, and at each timestep the environment returns the observation and a reward of $+1$. An episode is terminated after 200 time steps irrespective of the state of the cart. CartPole-v0 defines ``solving" as getting an average reward of at least 195.0 over 100 consecutive trials.

\paragraph{MountainCar.} This environment consists of an underpowered car on a one-dimensional track. At the beginning, the car is placed between two `hills'. The episode ends when the car reaches the top of the hill in front of it. Since the car is underpowered, the agent needs to drive it back and forth to build momentum. At every timestep the agent chooses to apply a force to move the car to the right, to the left, or to not apply a force. The goal is to end the episode in the fewest possible timesteps. 

We use the OpenAI Gym `MountainCar-v0' environment. This implementation is based on the system presented in~\citep{Moore-1991-15826}. The sensors provide the position and velocity of the car. The action space is discrete with three elements, and at each timestep the environment returns the observation and a reward of $-1$. An episode is terminated after 200 time steps irrespective of the state of the robot. MountainCar-v0  is considered ``solved" if the average reward over 100 consecutive trials is not less than -110.0.

\begin{figure*}[h]
	\vskip 0.1in
	\begin{equation*}
	\begin{array}{l}
0.97 * \pickc((0.0 - h_\tangle), -1) \hspace{-2pt}+ 0.05 * \foldc(+, (0.0 - h_\tangle))   \hspace{-2pt}+ 49.98 * (\pickc(h_\tangle, -2) - \pickc(h_\tangle, -1))
	\end{array}
	\end{equation*}
	\caption{A programmatic policy for steering, automatically discovered by the \algo algorithm with training on Aalborg. 
	}
	\label{fig:codesteer}
	\begin{equation*}
	\begin{array}{l }
	\ifc~(0.0001 - \pickc(h_\tangle, -1) > 0 )~\andc~(0.0001 + \pickc(h_\tangle, -1) > 0) \\
	\hspace{10pt} \thenc~ 0.95 * \pickc((0.64 - h_\rpm), -1)+ 5.02 * \foldc(+, (0.64 - h_\rpm))+ 43.89 * (\pickc(h_\rpm, -2) - \pickc(h_\rpm, -1)) \\
	\hspace{10pt} \elsec~ \;\, 0.95 * \pickc((0.60 - h_\rpm), -1)+ 5.02 * \foldc(+, (0.60 - h_\rpm))+ 43.89 * (\pickc(h_\rpm, -2) - \pickc(h_\rpm, -1))
	\end{array}
	\end{equation*}
	\caption{A programmatic policy for acceleration, automatically discovered by the \algo algorithm with training on CG-Speedway-1.}
	\label{fig:cgaccel}
	\begin{equation*}
	\begin{array}{l }
0.86 * \pickc((0.0 - h_\tangle), -1) \hspace{-2pt}+ 0.09 * \foldc(+, (0.0 - h_\tangle))   \hspace{-2pt}+ 46.51 * (\pickc(h_\tangle, -2) - \pickc(h_\tangle, -1))
	\end{array}
	\end{equation*}
	\caption{A programmatic policy for steering, automatically discovered by the \algo algorithm with training on CG-Speedway-1.}
	\label{fig:cgsteer}
\end{figure*}

\paragraph{Results.} Table~\ref{table:openai} shows rewards obtained by optimal policies found using various methods in these environments. The first row gives numbers for the \drl method. The rows \algo-SMT and \algo-BOPT  for versions of the \algo algorithm that respectively use SMT-based optimization and Bayesian optimization to find template parameters (more on this below).

\section{Additional Details on Algorithm}\label{sec:opt}

Now we elaborate on the optimization techniques we used in the distance computation step $\argmin_{e'} \sum_{h \in \mathcal{H}} \norm{e'(h) - e_\NN(h)}$, to find a program similar to a given program $e$, in Algorithm~1.

As mentioned in the main paper, we start by enumerating a list of {\em program templates}, or programs with numerical-valued parameters $\theta$. This is  done by first replacing the numerical constants in $e$ by parameters, eliding some subexpressions from the resulting parameterized program, and then regenerating the subexpressions using the rules of $\S$ (without instantiating the parameters), giving priority to shorter expressions. The resulting program template $e_\theta$ follows the sketch $\S$ and is also structurally close to $e$. Now we search for values for parameters $\theta$ that optimally imitate the neural oracle. 

\paragraph{Bayesian optimization.}
We use Bayesian optimization as our primary tool when searching for such optimal parameter values. This method applies to problems in which actions (program outputs) can be represented as 
vectors of real numbers. All problems considered in our experiments fall in this category. The distance of individual pairs of outputs of the synthesized program and the policy oracle is then simply the Euclidean distance between them. The sum of these distances is used to define the aggregate cost across all inputs in $\mathcal{H}$. We then use Bayesian optimization to find parameters that minimize this cost.

\begin{figure}[h]
	\begin{equation*}
	\begin{array}{l }
	\ifc~(0.1357 +\pickc(h_4, -1))< 0 \\
	\hspace{15pt}\thenc~2\\
	\hspace{15pt}\elsec~0
	\end{array}
	\end{equation*}
	\caption{A programmatic policy for Acrobot, automatically discovered by the \algo algorithm.}
	\label{fig:acro}
	
	\begin{equation*}
	\begin{array}{l }
	\ifc~(\foldc(+, h_0) - \pickc(h_3, -1)) > 0  \\
	\hspace{15pt}\thenc~0\\
	\hspace{15pt}\elsec~1
	\end{array}
	\end{equation*}
	\caption{A programmatic policy for CartPole, automatically discovered by the \algo algorithm.}
	\label{fig:cpole}

	\begin{equation*}
	\begin{array}{l }
	\ifc~(0.2498 - \pickc(h_0, -1) > 0 )\\ \hspace{15pt}\andc~(0.0035 - \pickc(h_1, -1) < 0) \\
	\hspace{30pt}\thenc~0\\
	\hspace{30pt}\elsec~2
	\end{array}
	\end{equation*}
	\caption{A programmatic policy for MountainCar, automatically discovered by the \algo algorithm.}
	\label{fig:mcar}

\end{figure}

\paragraph{SMT-based Optimization.}  
We also use a second parameter search technique based on SMT (Satisfiability Modulo Theories) solving.
Here, we generate a constraint that stipulates that for each $h \in \mathcal{H}$, the output $e_\theta(h)$ must match $e_\NN(h)$ up to a constant error. Here, $e_\NN(h)$ is a constant value obtained by executing $e_\NN$. The output $e_\theta(h)$ depends on unknown parameters $\theta$; however, constraints over $e_\theta(h)$ can be represented as constraints over $\theta$ using techniques for {\em symbolic execution} of programs~\citep{cadar2013symbolic}.
Because the oracle is only an approximation to the optimal policy in our setting, we do not insist that the generated constraint is satisfied entirely. Instead, we 
set up a Max-Sat problem which assigns a weight to the constraint for each input $h$, and then solve this problem with a Max-Sat solver. 

Unfortunately, SMT-based optimization does not scale well in environments with continuous actions. Consequently, we exclusively use Bayesian optimization for all \torcs based experiments. SMT-based optimization can be used in the classic control games, however, and Table~\ref{table:openai} shows results generated using this technique (in row \algo-SMT).

The results in Table~\ref{table:openai} show that for the classic control games, SMT-based optimization gives better results. This is because the small number of legal actions in these games, limited to at most three values $\{0,1,2\}$, are well suited for the SMT setting. The SMT solver is able to efficiently perform  parameter optimization, with a small set of histories. Whereas, the limited variability in actions forces the  Bayesian optimization method to use a larger set of histories, and makes it harder for the method to avoid getting trapped in local minimas. 

\section{Policy Examples}\label{sec:examples}
In this section we present more examples of the policies found by the \algo algorithm.

The program in Figure~\ref{fig:codesteer} shows the body of a policy for steering, which  together with the acceleration policy given in the paper (Figure~\ref{fig:code}), was found by the \algo algorithm  by training on the Aalborg track. Figures~\ref{fig:cgaccel} \& \ref{fig:cgsteer} likewise show the policies for acceleration and steering respectively, when trained on the CG-Speedway-1 track. Similarly, Figures~\ref{fig:acro}, \ref{fig:cpole} \& \ref{fig:mcar} show policies found for Acrobot, CartPole, and MountainCar respectively. Here $h_i$ is the sequence
of observations from the $i$-th of $k$ sensors, for example $h_0$ is the $0$-th sensor. The sensor order is determined by the OpenAI simulator.

\section{\torcs Video}
We provide a video at the following link, which depicts clips of the \drl agent and the \algo algorithm synthesized program, on the training track and one of the transfer (unseen) tracks, in that order:
 
\verb|https://goo.gl/Z2X5x6|

On the training track, we can see that the steering actions taken by the \drl agent are very irregular, especially when compared to the smooth steering actions of the \algo agent in the following clip. For the transfer track, we show the agents driving on the E-Road track. We can see that the \drl agent crashes before completing a full lap, while the \algo agent does not crash. We have provided only small clips of the car during a race, to keep the video length and size small, but the behavior is representative of the agent for the entire race.